\documentclass[10pt,twocolumn,letterpaper]{article}

\usepackage{iccv}
\usepackage{times}
\usepackage{epsfig}
\usepackage{graphicx}
\usepackage{amsmath}
\usepackage{amssymb}
\usepackage{booktabs}
\usepackage{multirow}
\usepackage[dvipsnames]{xcolor}
\usepackage{enumitem}
\usepackage{bbm}
\usepackage{caption}
\usepackage{subcaption}
\usepackage{pifont}
\newcommand{\cmark}{\ding{51}}%
\newcommand{\xmark}{\ding{55}}%
\usepackage[accsupp]{axessibility}

\usepackage[pagebackref=true,breaklinks=true,letterpaper=true,colorlinks,bookmarks=false]{hyperref}

\iccvfinalcopy 

\def\iccvPaperID{****} 

\ificcvfinal\fi

\begin{document}

\title{CheckerPose: Progressive Dense Keypoint Localization for Object Pose Estimation with Graph Neural Network}

\author{Ruyi Lian \quad Haibin Ling  \\
  {\normalsize Department of Computer Science,   Stony Brook University, Stony Brook, NY 11794-2424, USA} \\
  {\tt\small \{rulian,hling\}@cs.stonybrook.edu} \\
}

\maketitle
\ificcvfinal\thispagestyle{empty}\fi

\begin{abstract}
    Estimating the 6-DoF pose of a rigid object from a single RGB image is a crucial yet challenging task. Recent studies have shown the great potential of dense correspondence-based solutions, yet improvements are still needed to reach practical deployment. In this paper, we propose a novel pose estimation algorithm named \emph{CheckerPose}, which improves on three main aspects. Firstly, CheckerPose densely samples 3D keypoints from the surface of the 3D object and finds their 2D correspondences progressively in the 2D image. Compared to previous solutions that conduct dense sampling in the image space, our strategy enables the correspondence searching in a 2D grid (\ie, pixel coordinate). Secondly, for our 3D-to-2D correspondence, we design a compact binary code representation for 2D image locations. This representation not only allows for progressive correspondence refinement but also converts the correspondence regression to a more efficient classification problem. Thirdly, we adopt a graph neural network to explicitly model the interactions among the sampled 3D keypoints, further boosting the reliability and accuracy of the correspondences. Together, these novel components make CheckerPose a strong pose estimation algorithm. When evaluated on the popular Linemod, Linemod-O, and YCB-V object pose estimation benchmarks, CheckerPose clearly boosts the accuracy of correspondence-based methods and achieves state-of-the-art performances. {Code is available at} \url{https://github.com/RuyiLian/CheckerPose}.
\end{abstract}

\section{Introduction}
\label{sec:intro}

\begin{figure}[t]
\setlength{\abovecaptionskip}{-.05cm}
\begin{center}
   \includegraphics[width=0.95\linewidth]{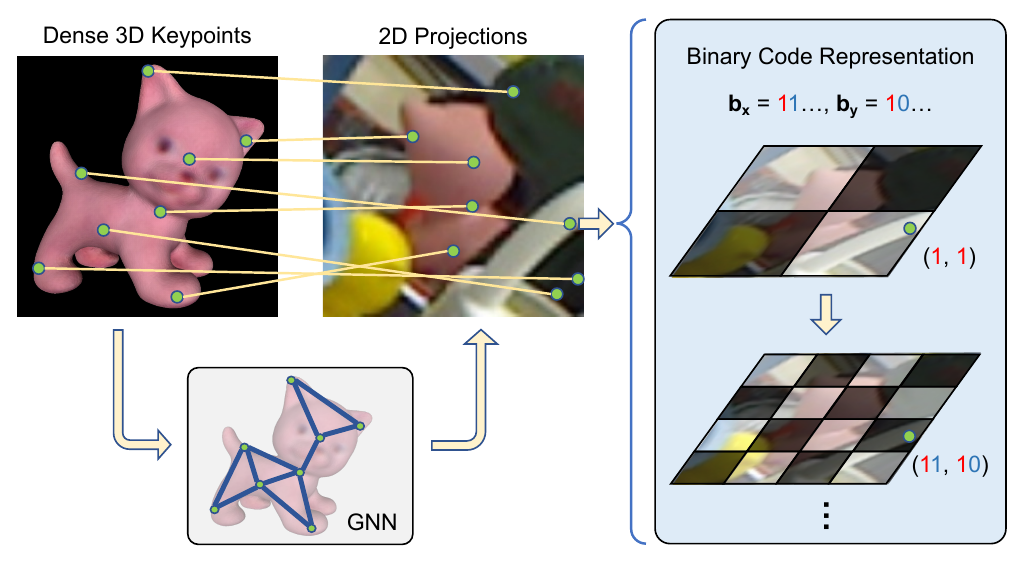}
\end{center}
   \caption{\textbf{Illustration of CheckerPose.} We evenly sample dense keypoints from the object surface, and predict the 2D locations in the input image. We design a binary code representation to progressively localize each keypoint in the iteratively refined 2D grids. To improve the localization, we also use graph neural networks to explicitly model the interactions between 3D keypoints. 
   Note: we plot 8 keypoints for better visualization, while use 512 keypoints in practice.}
\label{fig:motivation}
\end{figure}

Object pose estimation from RGB images aims to estimate the rotation and translation of a given rigid object relative to the camera. It is crucial in various applications including robot grasping and manipulation \cite{zhu2014single, tremblay2018deep, tremblay2020indirect}, autonomous driving \cite{manhardt2019roi, wu20196d, li2021exploring}, augmented reality \cite{marchand2015pose, tang20193d}, \etc. 
Most existing methods~\cite{rad2017bb8, tekin2018real, oberweger2018making, hu2019segmentation, peng2019pvnet, zakharov2019dpod, park2019pix2pose, li2019cdpn, song2020hybridpose} first estimate an intermediate geometric representation, \ie, the correspondences between 3D object keypoints and 2D image locations, and then recover the object pose using the Perspective-n-Point (PnP) algorithm. Theoretically, for a rigid object, four pairs of 3D-2D correspondences can determine a unique pose \cite{quan1999linear, gao2003complete, persson2018lambda}. In practice, however, sparse correspondences easily degrade due to occlusion, background clutter, lighting variation, \etc.

Increasing the number of 3D-2D correspondences is a feasible solution to enhance robustness, especially when combined with outlier removal mechanisms such as RANSAC. Recent methods~\cite{zakharov2019dpod, park2019pix2pose, li2019cdpn, hodan2020epos, hu2020single, wang2021gdr, di2021so} densely sample 2D image pixels and predict their 3D object coordinates. While these dense predictions improve the robustness of pose estimation, they have several drawbacks. Firstly, the predictions consider only visible pixels and ignore global relations between visible and occluded keypoints, making them unstable when the object is under severe occlusions. Secondly, estimating the corresponding 3D coordinates is nontrivial. Finally, the rich shape prior information is not effectively encoded.

To overcome the above issues, we propose a novel 6D pose estimation algorithm, named \emph{CheckerPose}, which improves dense correspondence with three cooperative components: dense 3D sampling, progressive 2D localization through binary coding, and shape prior encoding with graph neural network, as illustrated in Figure~\ref{fig:motivation}.

For dense correspondence, CheckerPose samples 3D keypoints on the object surface and then finds their 2D pixel correspondences in the 3D-to-2D matching way. Compared to previous solutions that conduct dense sampling in the 2D image space, our strategy enables more efficient correspondence searching in a 2D grid (\ie, pixel coordinate) using 2D binary coding, as well as explicit shape prior modeling with graph representation. 

Then, to facilitate the localization of dense keypoints, we propose a 2D hierarchical binary coding to represent a 2D image position. Specifically, we superpose a grid on the input image and predict which cells contain the desired keypoints. The precision of the 2D keypoint location is controlled by the resolution of the grid. This novel representation allows us to refine the correspondence progressively. We first localize the keypoints in the $2 \times 2$ grid, and then iteratively subdivide each cell and localize the keypoints in the refined grid. Inspired by ZebraPose~\cite{su2022zebrapose}, we use binary codes on the $x$ and $y$ directions to represent each cell, which makes the grids have a checkerboard pattern. 

Furthermore, to capture the shape prior of the 3D object, we adopt a graph neural network to explicitly model the interactions among the sampled 3D keypoints and to guide the progressive correspondence estimation. In particular, we construct the $k$-nearest neighbor ($k$-NN) graph of the dense keypoints and utilize graph network layers to fuse information from a keypoint and its neighbors. By stacking multiple such layers, we can capture non-local interactions between invisible and visible keypoints, and thus significantly improve the prediction robustness of invisible keypoints.

To summarize, our main contributions are as follows: 
\begin{itemize}[itemsep=0pt,topsep=0pt,parsep=0pt]
    \item We propose to localize dense 3D keypoints in the input image, to establish dense correspondences for instance-level object pose estimation.
    
    \item We design a hierarchical binary coding strategy for 2D projections, which enables progressive localization of dense keypoints.

    \item We utilize graph neural networks to explicitly model the interactions between 3D keypoints and improve the predictions of invisible keypoints.
\end{itemize}
Together, these novel contributions make our CheckerPose a strong pose estimation algorithm. We conduct extensive experiments on the popular benchmarks including Linemod \cite{hinterstoisser2012model}, Linemod-Occlusion \cite{brachmann2014learning}, and YCB-V \cite{xiang2017posecnn}, and CheckerPose consistently achieves state-of-the-art performances.

\begin{figure*}[t]
\setlength{\abovecaptionskip}{-.15cm}
\begin{center}
   \includegraphics[width=\linewidth]{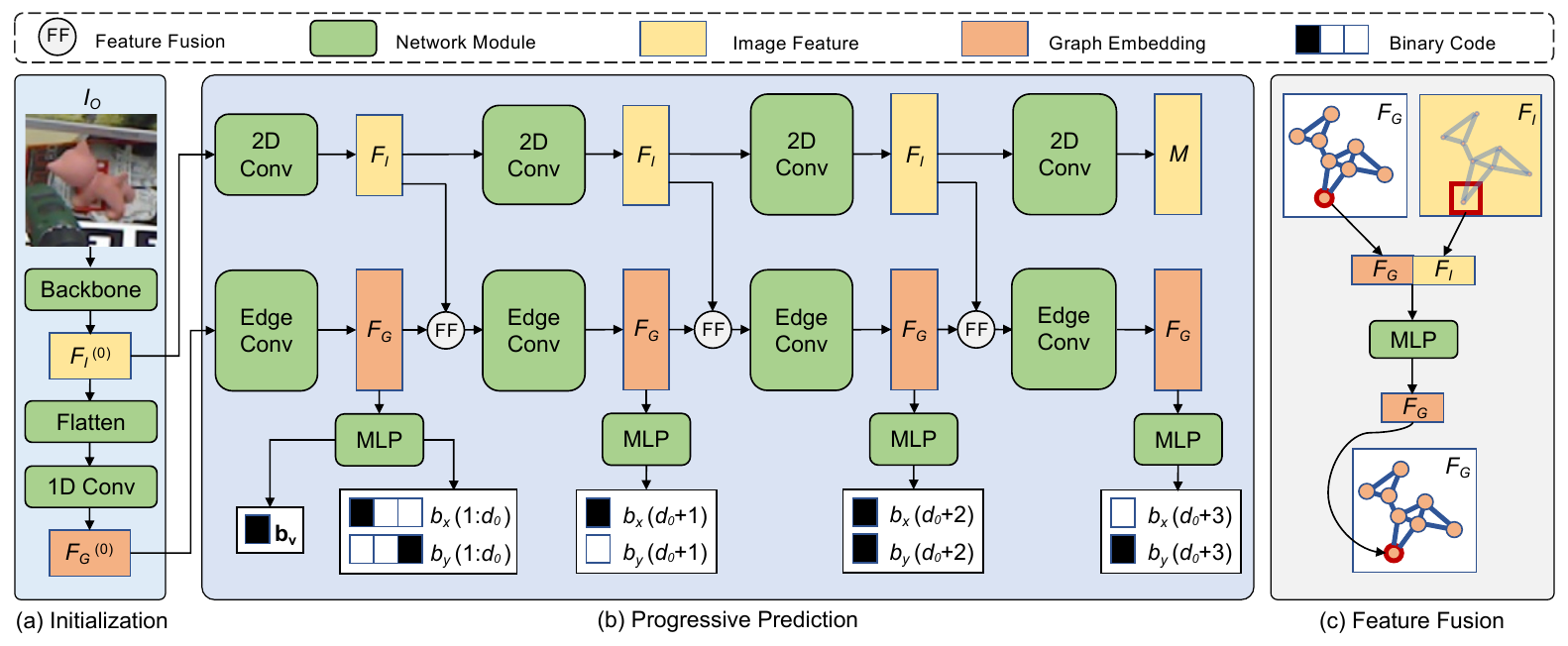}
\end{center}
   \caption{\textbf{Framework of our progressive dense keypoint localization with graph neural network, \ie, CheckerPose.}
   Given an RGB image and object detection results, we progressively generate the binary codes representing the 2D locations of $N$ 3D keypoints. \textbf{(a)} Initial graph embedding generation: we use a CNN backbone network to extract feature $F_I^{(0)}$ from the zoomed-in RoI $I_O$, and then transform $F_I^{(0)}$ to the initial keypoint embeddings $F_G^{(0)}$ in the $k$-NN graph $\mathcal{G}$. \textbf{(b)} Progressive prediction: we use a graph neural network to generate the binary code representation in a coarse-to-fine manner. We adopt an additional CNN decoder network to generate image features {with increased resolutions} from $F_I^{(0)}$, and fuse the features in the graph neural network based on current predictions. Object segmentation masks $M$ are predicted as an auxiliary learning task. {\textbf{(c)} Feature fusion: to fuse the image feature $F_I$ into the graph embeddings $F_G$, for each keypoint, we crop a feature patch from $F_I$ based on the current localization result, and concatenate the flattened feature with keypoint embedding. We then use a shared MLP to fuse the concatenation and the result is the updated keypoint embedding.}}
\label{fig:pipeline}
\end{figure*}

\section{Related Work}

In this section we review previous studies that are closely related to our work, mainly including different types of pose estimators and graph neural networks.

\vspace{-2.5mm}\paragraph{Direct Methods.} Given an input RGB image, direct methods estimate the 6D pose of the object in the image without intermediate geometric representations, \eg, 3D-2D correspondences. Traditional direct methods mainly adopt template matching techniques with hand-crafted features \cite{huttenlocher1993comparing, gu2010discriminative, hinterstoisser2011gradient}, and thus can not handle textureless objects well. Recent deep learning based methods utilize features learned by CNNs to directly regress 6D pose \cite{xiang2017posecnn} or formulate the rotation estimation as a classification task by discretizing the rotation space $SO(3)$ \cite{tulsiani2015viewpoints, su2015render, kehl2017ssd, sundermeyer2018implicit}.

\vspace{-2.5mm}\paragraph{Correspondence Guided Methods.} Instead of direct estimation, correspondence guided methods \cite{pavlakos20176, rad2017bb8, tekin2018real, oberweger2018making, hu2019segmentation, peng2019pvnet, hu2020single, hu2021wide, zakharov2019dpod, park2019pix2pose, li2019cdpn, wang2021gdr, di2021so, su2022zebrapose} follow a two-stage framework: they first predict a set of correspondences between 3D object frame coordinates and 2D image plane coordinates, and then recover the pose from the 3D-2D correspondences with a PnP algorithm \cite{lepetit2009epnp, kneip2014upnp, ferraz2014very, urban2016mlpnp, chen2020end}. RANSAC can be used to remove the outliers in the correspondences. 
Keypoint-localization based methods~\cite{pavlakos20176, rad2017bb8, tekin2018real, oberweger2018making, hu2019segmentation, peng2019pvnet, hu2020single, hu2021wide} estimate the 2D coordinates for a sparse set of predefined 3D keypoints, while dense methods~\cite{zakharov2019dpod, park2019pix2pose, li2019cdpn, wang2021gdr, di2021so, su2022zebrapose} predict the 3D object frame coordinate of each 2D image pixel. 
Compared with sparse correspondences, dense correspondences contain richer context information of the scene and is more robust to occlusion.

\vspace{-2.5mm}\paragraph{Graph Neural Networks for 3D Vision.} 
In 3D vision tasks, point clouds and meshes are important input data formats since they can efficiently represent complex shapes. Compared with convolutional neural networks (CNNs), graph neural networks (GNNs)~\cite{sperduti1997supervised} can handle inputs with irregular structures and effectively model the long-range dependencies, and thus are widely used for processing point clouds and meshes. While meshes can be naturally treated as graphs, a common practice of constructing graphs from point clouds is to treat each 3D point as graph nodes and connect each node to its $k$ nearest neighbors~\cite{wang2019dynamic, simonovsky2017dynamic, chen2020hierarchical}. GNN-based methods have been proposed for representation learning~\cite{simonovsky2017dynamic, wang2019dynamic, lin2020convolution, verma2018feastnet}, detection~\cite{shi2020point,chen2020hierarchical}, segmentation~\cite{qi20173d, li2019deepgcns}, data generation~\cite{qian2020pugeo, lin2021mesh}, camera pose inference~\cite{Li&Ling21iccv,Li&Ling22eccv}, \etc.
Graph techniques have also been used for learning dense correspondences between 3D shapes~\cite{saleh2022bending} using both local and global information. 
For object pose estimation, GNNs are mainly used for RGB-D inputs~\cite{chen2021fs, zhou2021pr} to enhance the feature extraction from different modalities. Another recent application is to learn geometric structures of the sparse keypoints for domain adaptation~\cite{zhang2021keypoint}. 

\vspace{1mm}\noindent\textbf{Our work} follows the two-stage framework and combines the strengths of both keypoint-based methods and dense methods, by localizing a dense set of predefined 3D keypoints to establish dense correspondences. Moreover, it utilizes GNNs to efficiently model the interactions among dense 3D keypoints and thus improve the localization in the input RGB image for monocular object pose estimation.

\section{Method}

\subsection{Problem Formulation and Method Overview}

Given an RGB image $I$ and a rigid object $O$, our goal is to estimate rotation $\mathbf{R} \in SO(3)$ and translation $\mathbf{t} \in \mathbb{R}^3$ of $O$ relative to the calibrated camera. We assume the 3D geometry information, \eg, the 3D CAD model, is available, thus we can obtain $N (N\gg 8)$ keypoints $\mathcal{P}\subset\mathbb{R}^3$ from the object surface using farthest point sampling (FPS).

We adopt a two-stage pipeline for object pose estimation: we first predict 2D projection $\boldsymbol\rho \in \mathbb{R}^2$ for each keypoint $P \in \mathcal{P}$, and then regress the rotation and translation from the 3D-2D correspondences via a PnP solver. 
For the input RGB image, we use an off-the-shelf object detector~\cite{ren2015faster, tian2019fcos} to detect the object bounding box and extract the zoomed-in Region of Interest (RoI) $I_O$, following the common practice in instance-level object pose estimation~\cite{li2019cdpn, wang2021gdr, di2021so, su2022zebrapose}. 
Figure~\ref{fig:pipeline} illustrates our proposed pipeline.
We first process the input RoI $I_O$ by a backbone network to obtain backbone feature $F_I^{(0)}$ and keypoint embedding $F_G^{(0)}$ in the $k$-NN graph $\mathcal{G}$. Then we use graph network layers (\ie, EdgeConv~\cite{wang2019dynamic}) to progressively localize the keypoints, which are represented as binary codes $\mathbf{b_v}, \mathbf{b_x}$, and $\mathbf{b_y}$. We also use a standard CNN decoder to transform $F_I^{(0)}$ to a series of image feature maps, and fuse the features in the graph neural network based on the current predicted locations. The CNN decoder also outputs object segmentation masks $M$ as an auxiliary learning task. Finally, we convert the binary codes to 2D coordinates and use a PnP solver to recover the pose from the established correspondences.
We describe our method, named \textit{CheckerPose} due to the checkerboard-like binary pattern, in details as follows.

\subsection{Hierarchical Representation of 2D Keypoints}
\label{sec:binary_code_repr}

\begin{figure}[t]
\setlength{\abovecaptionskip}{-.1cm}
\begin{center}
   \includegraphics[width=0.95\linewidth]{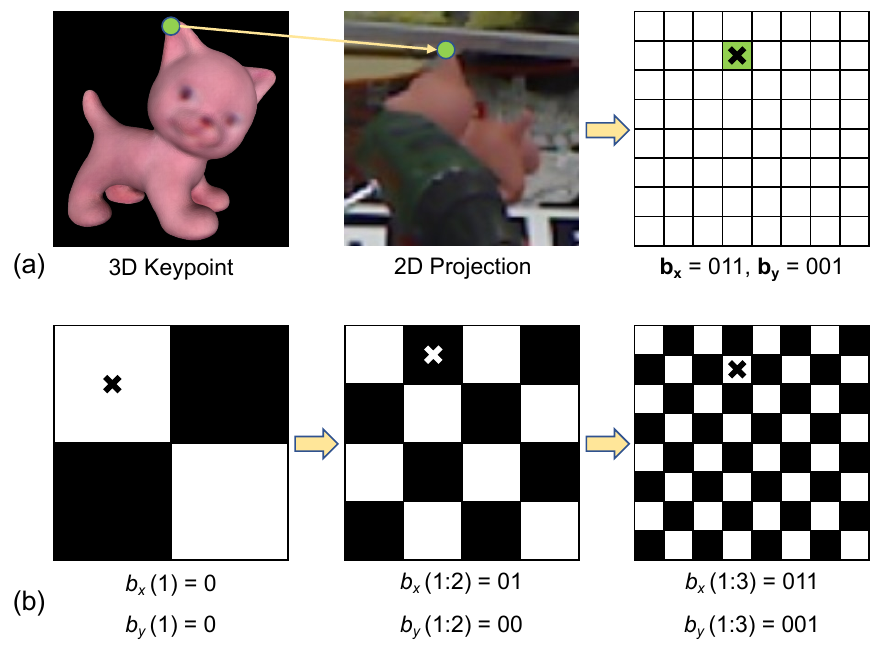}
\end{center}
   \caption{\textbf{Keypoint location representation.} \textbf{(a)} We represent the 2D projection coordinate as the center of the cell containing the 2D projection. \textbf{(b)} We iteratively refine the grid and represent the cell as binary codes $\mathbf{b_x}, \mathbf{b_y}$.}
\label{fig:representation}
\end{figure}

Establishing 3D-2D correspondences provides an intermediate representation for object pose estimation. In this work, we focus on localizing a dense set of predefined 3D keypoints $\mathcal{P}$ in the 2D image plane. 
For $N (N \gg 8)$ 3D keypoints $\mathcal{P}$, we first predict whether their 2D projections appear in the RoI $I_O$, and then localize the keypoints inside $I_O$, denoted as $\mathcal{P}_I$. In contrast to directly regressing the precise coordinates, we superpose a $2^d \times 2^d$ grid $S$ on the RoI $I_O$ and predict which cell $s \in S$ contains the 2D projection $\boldsymbol\rho$ (Figure~\ref{fig:representation} (a)). Then we can use the coordinate of the cell center to approximate $\boldsymbol\rho$, and only need to predict the discrete index $(i_x, i_y) (0 \leq i_x, i_y \leq 2^d-1)$ of the cell $s$, which is much easier than precise regression. The localization precision is controlled by the resolution of the grid $S$, and approaches the actual 2D projection as $d \rightarrow \infty$.

Based on the approximate representation, we can further localize the keypoint $P \in \mathcal{P}_I$ in a coarse-to-fine manner. As shown in Figure~\ref{fig:representation} (b), at the beginning, we superpose a $2 \times 2$ grid $S^{(1)}$ on the RoI $I_O$ and predict the index of the cell $s_P^{(1)}$. Then at iteration $j\ (2 \leq j \leq d)$, we increase the grid resolution from $2^{j-1} \times 2^{j-1}$ to $2^{j} \times 2^{j}$ by evenly splitting each cell $s^{(j-1)} \in S^{(j-1)}$ into halves on both $x$ and $y$ directions. With the prediction of $s_P^{(j-1)}$ in iteration $j-1$, we only need to search the corresponding $2 \times 2$ sub-cells to find $s_P^{(j)}$ in the refined grid $S^{(j)}$.  

Inspired by ZebraPose~\cite{su2022zebrapose}, we use binary codes to concisely represent the hierarchical localization. For the cell $s_P$ in the final $2^d \times 2^d$ grid $S$, we use a $d$-bit binary code $\mathbf{b_x}$ to represent the index $i_x$ as
\begin{equation}
    i_x = \sum_{k=1}^d b_x(k) \times 2^{d-k},
    \label{eq:code2index}
\end{equation}
where $b_x(k)$ is the $k$-th bit of $\mathbf{b_x}$. We use another $d$-bit binary code $\mathbf{b_y}$ to represent the index $i_y$ in the same way.
The first $j (1 \leq j \leq d)$ bits of $\mathbf{b_x}$ and $\mathbf{b_y}$ also represent the cell $s_P^{(j)} \in S^{(j)}$.
We use an additional $1$-bit binary code $\mathbf{b_v}$ to indicate the existence of the projection $\boldsymbol\rho$ in the RoI $I_O$, where $\mathbf{b_v} = 1$ means $\boldsymbol\rho \in I_O$ while $\mathbf{b_v} = 0$ means $\boldsymbol\rho \notin I_O$. 

Compared with dense representations (\eg, heatmaps~\cite{pavlakos20176, oberweger2018making} and vector-fields~\cite{peng2019pvnet, hu2019segmentation}), our representation needs only $2d+1$ binary bits for each keypoint, thus greatly reduces the memory usage for dense keypoint localization. 
In addition, during inference, we can efficiently convert the binary codes to the 2D coordinates. 
Furthermore, our representation can be naturally predicted in a progressive way, which allows to gradually improve the localization via iterative refinements.

\subsection{Dense Keypoint Localization via Graph Neural Network}

Modeling the interactions among the keypoints $\mathcal{P}$ is crucial for predicting their 2D locations. For the keypoints that are invisible due to occlusions or self-occlusions, the features of the visible ones provide additional clues to infer the 2D locations. 
However, previous keypoint-based methods mainly use convolutional neural networks (CNNs), which can not handle inputs with irregular structure and thus fail to explicitly capture the interactions among $\mathcal{P}$. 

We instead utilize graph neural networks (GNNs) to process the features $F = \{f_1, \cdots, f_N \}$ of $N$ keypoints $\mathcal{P}$. 
To construct a graph $\mathcal{G}$ from $\mathcal{P}$, we treat each keypoint $P_i \in \mathcal{P} (1 \leq i \leq N)$ as a graph node, and connect $P_i$ to its $k$ nearest neighbors in 3D Euclidean space to generate edges $\mathcal{E}$. 
We adopt the EdgeConv operation~\cite{wang2019dynamic} as our graph network layer, which directly models local interactions between $P_i$ and its neighbors. For edge $(i, j) \in \mathcal{E}$, we compute the feature $e_{ij}$ as 
\begin{equation}
    e_{ijm} = \mathrm{ReLU} (\theta_m \cdot (f_j - f_i) + \phi_m \cdot f_i),
    \label{eq:edge_feature}
\end{equation}
where $e_{ijm}$ is the $m$-th channel of $e_{ij}$, and $\theta_m, \phi_m$ are the weights of the filters. The feature of $P_i$ is updated by aggregating the edge features as
\begin{equation}
    f_{im}^{'} = \max_{j: (i, j) \in \mathcal{E}} e_{ijm},
    \label{eq:edge_conv_aggregate}
\end{equation}
where $f_{im}^{'}$ is the $m$-th channel of updated feature $f_{i}^{'}$. 
By stacking multiple EdgeConv operations, our network can gradually learn the non-local interactions in a computationally efficient way for dense keypoints $\mathcal{P}$.

As shown in Figure~\ref{fig:pipeline} (a), to obtain the initial keypoint embeddings $F_G^{(0)}$ in $\mathcal{G}$, we first use a backbone network to extract a $C_0 \times 2^{d_0} \times 2^{d_0}$ feature map $F_I^{(0)}$ from RoI $I_O$, where $C_0$ is the number of the feature channels, and $2^{d_0} \times 2^{d_0}$ is the spatial size. We then reshape $F_I^{(0)}$ to $C_0 \times 2^{2d_0}$ by flattening the spatial dimensions, and use a 1D convolutional network layer to obtain a $N \times 2^{2d_0}$ feature map, which is regarded as the initial $2^{2d_0}$-dimensional embeddings $F_G^{(0)}$ for $N$ keypoints. 

After obtaining $F_G^{(0)}$, we use a graph neural network to predict the 1-bit indicator code $\mathbf{b_v}$, and progressively generate the $d$-bit index codes $\mathbf{b_x}, \mathbf{b_y}$. Specifically, at stage 0, we apply $L_0$ EdgeConv~\cite{wang2019dynamic} operations to $F_G^{(0)}$ to get the updated embeddings $F_G^{(1)}$, and then use shared MLPs to generate $\mathbf{b_v}$ and the first $d_0$ bits of $\mathbf{b_x}, \mathbf{b_y}$, respectively. Then at stage $j (1 \leq j \leq d - d_0)$, we apply $L_j$ EdgeConv operations to $F_G^{(j)}$ to obtain $F_G^{(j+1)}$, and use shared MLPs to generate new bits $b_x(d_0 + j), b_y(d_0 + j)$ for $\mathbf{b_x}, \mathbf{b_y}$, respectively. We regard stage $j (1 \leq j \leq d - d_0)$ as refinement stage, since it refines the localization from the low-resolution grid $S^{(d_0 + j - 1)}$ to the high-resolution one $S^{(d_0 + j)}$.

Compared with generating all bits at the network output layer, our progressive prediction enables image feature fusion at each refinement stage. 
As shown in Figure~\ref{fig:pipeline} (b), starting with the image feature map $F_I^{(0)}$ with low spatial resolution $2^{d_0} \times 2^{d_0}$, we use an additional CNN-based decoder to progressively generate image feature maps $F_I^{(1)}, \cdots, F_I^{(d - d_0)}$ with increased spatial resolutions $2^{d_0 + 1} \times 2^{d_0 + 1}, \cdots, 2^{d} \times 2^{d}$, respectively. We also add skip connections between the backbone and the decoder to {recover the high-resolution details lost in $F_I^{(0)}$.}
{As shown in Figure~\ref{fig:pipeline} (c), at the beginning of the refinement stage $j$, for each keypoint $P$, we select local image feature from $F_I^{(j)}$ based on the localization result in the previous stage. We then concatenate $F_l^{(j)}$ with the keypoint embedding in the graph $\mathcal{G}$, and use a shared MLP to fuse the concatenation. The fused feature is used as the updated keypoint embedding.}
Since the initial keypoint embeddings $F_G^{(0)}$ are obtained from $F_I^{(0)}$, fusing the local image features in the refinement stages provides critical {high-resolution details} for fine-grained localization.

\subsection{Training}

For the 1-bit indicator code $\mathbf{b_v}$ of keypoint $P \in \mathcal{P}$, our network output $\hat{\mathbf{b}}_{\mathbf{v}}$ is the probability that $\mathbf{b_v} = 1$. We use binary cross-entropy loss for $\mathbf{b_v}$ as below:
\begin{equation}
    \mathcal{L}_v = \frac{1}{N} \sum_{P \in \mathcal{P}} \mathbf{b_v} \log \hat{\mathbf{b}}_{\mathbf{v}} + (1 - \mathbf{b_v}) \log(1 - \hat{\mathbf{b}}_{\mathbf{v}}),
    \label{eq:indicator_code_loss}
\end{equation}
where $N$ is the number of the keypoints. For $d$-bit index codes $\mathbf{b_x}, \mathbf{b_y}$, since we only localize the keypoints inside the RoI (\ie, $\mathbf{b_v} = 1$), denoted as $\mathcal{P}_I$, we compute binary cross-entropy loss for each bit of $\mathbf{b_x}$ as
\begin{multline}
    \mathcal{L}_x = \frac{1}{d N_I} \sum_{P \in \mathcal{P}_I} 
    \sum_{k=1}^d b_x(k) \log (\hat{b}_x(k)) + \\ (1 - b_x(k)) \log(1 - \hat{b}_x(k)),
    \label{eq:index_code_loss}
\end{multline}
where $N_I$ is the number of keypoints inside the RoI, $\hat{b}_x(k)$ is the network prediction for $k$-th bit of $\mathbf{b_x}$. We compute the loss $\mathcal{L}_y$ for $\mathbf{b_y}$ in the same way as $\mathcal{L}_x$.  

Besides predicting the 2D projections as binary codes, we also enforce the network to output object segmentation masks. To do this, we apply a single CNN layer to the final image feature map $F_I^{(d - d_0)}$ and obtain a $2 \times 2^d \times 2^d$ output, which serves as the full segmentation mask $M_{\mathrm{full}}$ and the visible one $M_{\mathrm{vis}}$. We input the network predictions to the sigmoid function and apply $L_1$ loss as the mask loss $\mathcal{L}_{\mathrm{mask}}$. Generating the masks can be regarded as an auxiliary task to facilitate the learning of image features. 

The overall loss function $\mathcal{L}$ is a combination of $\mathcal{L}_v$, $\mathcal{L}_x$, $\mathcal{L}_y$, and $\mathcal{L}_{\mathrm{mask}}$ as
\begin{equation}
    \mathcal{L} = \mathcal{L}_v + \mathcal{L}_x + \mathcal{L}_y + \mathcal{L}_{\mathrm{mask}}.
    \label{eq:loss_weighted}
\end{equation}

Before training the whole network, we pretrain the layers that generate $\mathbf{b_v}$ and the first $d_0$ bits of $\mathbf{b_x}, \mathbf{b_y}$. This encourages the backbone network to quickly adapt to the object keypoints with smaller GPU memory usage, and makes the initial localization to be good for local image feature fusion in the refinement stages. 

\subsection{Inference}
\label{sec:inference}

During inference, we first discard the keypoints with $\mathbf{b_v} = 0$. 
Then we convert the binary codes to the corresponding cells in the final grid $S$ (Eq.~\ref{eq:code2index}), and use the 2D coordinates of the cell centers as the keypoint projections. In this way, we establish dense 3D-2D correspondences from the network outputs without time-consuming computation operations, \eg, voting for the vector-field representations~\cite{peng2019pvnet}. Finally we use the RANSAC/PnP~\cite{lepetit2009epnp} or Progressive-X~\cite{barath2019progressive} solvers to obtain the object pose from the dense 3D-2D correspondences.
 
We empirically find that for textureless objects with severe self-occlusions, discarding the correspondences outside $M_{\mathrm{vis}}$ can improve the pose estimation results. 
To quantify the self-occlusions of a given object $O$, we uniformly sample 2,562 camera viewpoints on a sphere, and use the Hidden Point Removal (HPR) operator~\cite{katz2007direct} to estimate the visibility of point $P \in O$ from each viewpoint. We then calculate the proportion of the viewpoints for which $P$ is visible, denoted as $V(P)$. If $0.2 \leq V(P) < 0.4$, then $P$ is considered to be easily self-occluded. Note we ignore the points with $V(P) < 0.2$, to make our estimation robust to the classification error of the HPR operator. The overall self-occlusion of the object $O$ can be computed by
\begin{equation}
    r_{\rm so}(O) = \frac{1}{|O|} \sum_{P \in O} \mathbbm{1}(0.2 \leq V(P) < 0.4), 
    \label{eq:overall_selfocc}
\end{equation}
where $|O|$ is the number of vertices of the object CAD model, and $\mathbbm{1}(\cdot)$ is the indicator function. If $r_{\rm so}(O) \geq 0.5$, \ie, over half part of $O$ is easily to be self-occluded, then we regard $O$ as severely self-occluded.

\section{Experiments}

\subsection{Experimental Setup}
\paragraph{Implementation Details.} Our method is implemented using PyTorch~\cite{paszke2019pytorch} and trained using the Adam optimizer~\cite{kingma2014adam} with a batch size of 32. We pretrain our network for $50,000$ steps with learning rate of 2e-4. 
We use $N = 512$ keypoints, and utilize $k = 20$ nearest neighbors to construct the $k$-NN graph $\mathcal{G}$. 
For the binary code representation, we set $d = 6$ and $d_0 = 3$. We resize the input RoIs to $256 \times 256$, and use HRNet~\cite{wang2020deep} as our image feature backbone to extract $1024 \times 8 \times 8$ feature map $F_I^{(0)}$. Then we apply $L_0 = 2$ EdgeConv operations to get $\mathbf{b_v}$ and the first $d_0 = 3$ bits of $\mathbf{b_x}, \mathbf{b_y}$, and obtain the full binary codes after 3 refinement stages with $L_j = 3\ (j = 1, 2, 3)$ EdgeConv operations. 

\vspace{-2.5mm}\paragraph{Datasets.} We conduct our experiments on three commonly-used datasets for object pose estimation: Linemod (LM) \cite{hinterstoisser2012model}, Linemod-Occlusion (LM-O) \cite{brachmann2014learning}, and YCB-V \cite{xiang2017posecnn}.
LM consists of 13 sequences of real images with ground truth poses for a single object with background clutter and mild occlusion. Each sequence contains around $1,200$ images. Following \cite{brachmann2016uncertainty}, we utilize about $15\%$ images for training while keeping the rest for testing. We additionally use $1,000$ synthetic RGB images for each object during training following \cite{li2019cdpn, wang2021gdr, di2021so}. 
LM-O consists of $1,214$ images from a sequence of LM~\cite{hinterstoisser2012model}, where ground truth poses of eight objects with partial occlusion are annotated for testing. YCB-V is composed of more than $110,000$ real images of $21$ objects with severe occlusion and clutter. Apart from the real training images, we also utilize the physically-based rendered data following \cite{hodan2020bop} for training on LM-O and YCB-V.

\vspace{-2.5mm}\paragraph{Evaluation Metrics.} We employ the common evaluation metric ADD(-S) for object pose estimation. ADD(-S) measures whether the average distance between the model points transformed by the predicted pose and the ground truth is less than $10\%$ of the object's diameter (0.1d). For symmetric objects, ADD(-S) metric computes the deviation to the closest model point. On YCB-V, we also compute the AUC (area under curve) of ADD-S and ADD(-S) with a maximum threshold of 10 cm \cite{xiang2017posecnn}. On LM, we also report the $n^\circ, n$ cm metric, measuring the percentage of predicted 6D poses with rotation error below $n^\circ$ and translation error below $n$ cm. For symmetric objects $n^\circ, n$ cm computes the smallest error for all possible ground truth poses~\cite{li2018deepim, wang2021gdr}.

\begin{table}
  \centering
  \small
  \renewcommand{\tabcolsep}{1.5mm}
\begin{tabular}{ c | c | c | c | c | c }
\toprule
\multirow{2}{*}{Method} & \multicolumn{3}{c|}{ADD(-S)} & \multirow{2}{*}{$2^\circ2$cm} & \multirow{2}{*}{$5^\circ5$cm} \\
\cline{2-4}
 & 0.02d  &  0.05d  &  0.1d &  &   \\
\midrule
GDR-Net~\cite{wang2021gdr}   & 35.5 & 76.3 & 93.7 & 62.1 & N/A \\
SO-Pose~\cite{di2021so}      & \textbf{45.9}	& 83.1 & 96.0	& 76.9 & 98.5 \\
EPro-PnP~\cite{chen2022epro} & 44.8 &	82.0 & 95.8	& \textbf{81.0} & 98.5 \\
\midrule
Ours (w/o GNN) & 26.4 &	77.8 & 95.2 &	67.7 & 97.9 \\
Ours (w/o Prog.) & 14.1 & 56.9 & 85.8 &	42.3 & 94.1 \\
Ours (w/o $M_{\mathrm{full}}$) & 30.2 &	82.8 & 96.7 & 79.3 & \textbf{98.9} \\
Ours (w/o $M_{\mathrm{vis}}$) & 34.1 & 82.8 & 96.6 & 79.1 & \textbf{98.9} \\
\midrule
Ours (ResNet34) & 31.3 & 80.2 &	95.6 & 74.2 & 98.6 \\
Ours (RANSAC/PnP) & 31.1 & 81.4 & 96.6 & 78.4 &	\textbf{98.9} \\
\midrule
CheckerPose (Ours)                & 35.7 & \textbf{84.5} & \textbf{97.1} & 79.7 & \textbf{98.9} \\
\bottomrule
\end{tabular}
  \caption{\textbf{Ablation Study on the LM Dataset.} 
  }
  \label{tab:lm_ablation}
\end{table}

\begin{table*}
  \centering
  \small
\begin{tabular}{c|c|c|c|c|c|c|c|c}
\toprule
Method & PVNet~\cite{peng2019pvnet} & S.~Stage~\cite{hu2020single} & Hybrid~\cite{song2020hybridpose} & RePose~\cite{iwase2021repose} & GDR-Net~\cite{wang2021gdr} & SO-Pose~\cite{di2021so} & Zebra~\cite{su2022zebrapose}  & Ours \\
\midrule
ape     & 15.8 & 19.2 & 20.9 & 31.1 & 46.8 & 48.4 & \textcolor{blue}{57.9} & \textcolor{red}{58.3} \\
can     & 63.3 & 65.1 & 75.3 & 80.0 & 90.8 & 85.8 & \textcolor{blue}{95.0} & \textcolor{red}{95.7} \\
cat     & 16.7 & 18.9 & 24.9 & 25.6 & 40.5 & 32.7 &\textcolor{blue}{60.6} & \textcolor{red}{62.3} \\
driller & 65.7 & 69.0 & 70.2 & 73.1 & 82.6 & 77.4 & \textcolor{red}{94.8} & \textcolor{blue}{93.7} \\
duck    & 25.2 & 25.3 & 27.9 & 43.0 & 46.9 & 48.9 &\textcolor{blue}{64.5} & \textcolor{red}{69.9} \\
eggbox* & 50.2 & 52.0 & 52.4 & 51.7 & 54.2 & 52.4 & \textcolor{red}{70.9} & \textcolor{blue}{70.0} \\
glue*   & 49.6 & 51.4 & 53.8 & 54.3 & 75.8 & 78.3 & \textcolor{red}{88.7} & \textcolor{blue}{86.4} \\
holep.  & 36.1 & 45.6 & 54.2 & 53.6 & 60.1 & 75.3 & \textcolor{blue}{83.0} & \textcolor{red}{83.8} \\
\midrule
mean & 40.8 & 43.3 & 47.5 & 51.6 & 62.2 & 62.3 & \textcolor{blue}{76.9} & \textcolor{red}{77.5} \\
\bottomrule
\end{tabular}
  \caption{\textbf{Comparison with State-of-the-art Methods on the LM-O Dataset.} We report the Average Recall (\%) of ADD(-S). (*) denotes symmetric objects. We highlight the best result in red color, and the second best result in blue color.}
  \label{tab:result_lmo}
\end{table*}

\subsection{Ablation Study on LINEMOD Dataset}

We present ablation experiments on LM~\cite{hinterstoisser2012model} in Table~\ref{tab:lm_ablation} to verify the effectiveness of each module. 
{We also study the number of keypoint $N$ and the size of neighborhood $k$ in Supplementary.} 
We train a single pose estimator for all objects for 120k steps, with a fixed learning rate of 1e-4 for the first 100k steps and a smaller learning rate of 5e-5 for the remaining steps. During inference, we utilize the detection results from Faster-RCNN \cite{ren2015faster} by \cite{li2019cdpn}. We do not use any segmentation masks to filter the correspondences for fair comparison. Without specification, we use Progressive-X~\cite{barath2019progressive} to compute pose from the dense correspondences. 

\vspace{-2.5mm}\paragraph{Comparison with State of the Art.} As shown in Table~\ref{tab:lm_ablation}, our method outperforms the state-of-the-art methods~\cite{wang2021gdr, di2021so, chen2022epro} w.r.t.~ADD(-S) 0.05d, ADD(-S) 0.1d,\ and $5^\circ5$cm, and achieves comparable results w.r.t.~ADD(-S) 0.02d and $2^\circ2$cm. The improvement of ADD(-S) 0.1d indicates that our method can facilitate the estimation of hard cases and serve as a good initialization for refinement methods~\cite{li2018deepim, iwase2021repose, xu2022rnnpose}. Since the 2D coordinates of our estimated correspondences are approximated by the cell centers (Sec.~\ref{sec:binary_code_repr}), our pose estimation results in terms of ADD(-S) 0.02d may be further improved by increasing the grid resolution. 

\vspace{-2.5mm}\paragraph{Effectiveness of Graph Neural Networks.} Our network utilizes GNN layers, \eg, EdgeConv~\cite{wang2019dynamic}, to explicitly model the interactions between different keypoints. We also report the result of removing all GNN layers in Table~\ref{tab:lm_ablation}. Without GNN layers, the keypoints still interact indirectly via local image feature fusion modules, since the keypoints with close 2D locations share the similar local image features. However, the performance of pose estimation degrades significantly, demonstrating that it is important to directly model the keypoint interactions with GNN layers.

\vspace{-2.5mm}\paragraph{Effectiveness of Progressive Prediction.} Progressively generating the binary codes enforces our network to gradually refine the localization in the iteratively subdivided grids. It also enables image feature fusion based on the intermediate estimations, which can provide crucial {high-resolution details} for fine-grained localization. As shown in Table~\ref{tab:lm_ablation}, the accuracy decreases significantly without progressively generating the binary codes, which clearly demonstrates the importance of progressive prediction.

\vspace{-2.5mm}\paragraph{Effectiveness of Object Segmentation Masks.} Our network outputs the full segmentation mask $M_{\mathrm{full}}$ and the visible one $M_{\mathrm{vis}}$ as auxiliary tasks. As shown in Table~\ref{tab:lm_ablation}, the performance degrades without either $M_{\mathrm{full}}$ or $M_{\mathrm{vis}}$. 
The ADD(-S) 0.02d metric drops significantly without $M_{\mathrm{full}}$, indicating that predicting $M_{\mathrm{full}}$ facilitates image feature extraction for keypoint localization, since all the keypoints should be located within $M_{\mathrm{full}}$. The degraded performance without $M_{\mathrm{vis}}$ also implies that predicting $M_{\mathrm{vis}}$ provides important context information including occlusions.

\vspace{-2.5mm}\paragraph{Impact of Backbone Networks.} We report the results of our method with different backbone networks in Table~\ref{tab:lm_ablation}. After replacing HRNet~\cite{wang2020deep} by ResNet34~\cite{he2016deep}, our method still achieves comparable results with state of the art, which demonstrates the efficacy of our method regardless of the backbone networks.

\vspace{-2.5mm}\paragraph{Influence of PnP Solvers.} We show the results with different PnP solvers during inference in Table~\ref{tab:lm_ablation}. Since our correspondences are established from the binary codes, a small perturbation of our network prediction can result in flipped bit values, which may correspond to dramatically different locations in the input RoI. Compared with RANSAC/PnP~\cite{lepetit2009epnp}, Progressive-X~\cite{barath2019progressive} contains a spatial coherence filter to efficiently remove such outliers, and thus achieves better performance w.r.t.~to all the metrics, especially ADD(-S) 0.02d.

\subsection{Comparison to State of the Art}
In this section we present the quantitative results of our method on LM-O and YCB-V datasets. We train a single CheckerPose for each object for 380,000 steps with a fixed learning rate of 1e-4. During inference, we utilize the detections from FCOS \cite{tian2019fcos} provided by CDPNv2~\cite{li2019cdpn}. 

\vspace{-4mm}\paragraph{Experiments on the LM-O dataset.} We report the recall of ADD(-S) metric for the LM-O dataset in Table~\ref{tab:result_lmo}. 
Based on the criterion discussed in Sec.~\ref{sec:inference}, we filter out the correspondences outside the visible segmentation masks $M_{\mathrm{vis}}$ for textureless objects with severe self-occlusions, including can, cat, driller, and eggbox. 
Without the filtering operation, the average recall of ADD(-S) of our method is 77.1, which surpasses previous methods. The detailed results of each object without filtering are provided in supplementary material.
The additional filtering operation further improves the performance of our method. The intuition is that it is infrequent to observe an easily self-occluded keypoint $P$ in the training images. Besides, due to the lack of texture, it is also hard to infer the location of $P$ from other keypoints with distinguishable features. Such objects may require much more training steps to achieve stable estimations for easily self-occluded keypoints. Simply discarding correspondences outside $M_{\mathrm{vis}}$ reduces unstable localization results when our network is trained for limited steps, and enhances the robustness of pose estimation.

\vspace{-4.5mm}
\paragraph{Experiments on the YCB-Video dataset.} We report the averaged metrics of 21 objects in Table~\ref{tab:result_ycbv}, and provide detailed results in the suppl.. Based on the criterion discussed in Sec.~\ref{sec:inference}, we use visible segmentation masks to filter correspondences for foam\_brick. We also apply the filtering operation to pudding\_box because it is severely occluded by gelatin\_box, which is a distraction object with similar texture. As shown in Table~\ref{tab:result_ycbv}, CheckerPose achieves the best performance w.r.t. ADD(-S) and AUC of ADD(-S), and is comparable with state of the art w.r.t. AUC of ADD-S.

\begin{table}
  \centering
\begin{tabular}{l|c|c|c}
\toprule
\multirow{2}{*}{Method} & \multirow{2}{*}{ADD(-S)} & AUC & AUC \\
 &  & ADD-S & ADD(-S) \\
\midrule
SegDriven~\cite{hu2019segmentation} & 39.0 & -- & -- \\
SingleStage~\cite{hu2020single}     & 53.9 & -- & -- \\
CosyPose~\cite{labbe2020cosypose}   & --  & 89.8 & 84.5 \\
RePose~\cite{iwase2021repose}       & 62.1 & 88.5 & 82.0 \\
GDR-Net~\cite{wang2021gdr}              & 60.1 & \textcolor{red}{91.6} & 84.4 \\
SO-Pose~\cite{di2021so}             & 56.8 & 90.9 & 83.9 \\
ZebraPose~\cite{su2022zebrapose}        & \textcolor{blue}{80.5} & 90.1 & \textcolor{blue}{85.3} \\
DProST~\cite{park2021dprost}        & 65.1 & -- & 77.4 \\
CheckerPose (Ours)                                & \textcolor{red}{81.4} & \textcolor{blue}{91.3} & \textcolor{red}{86.4} \\
\bottomrule
\end{tabular}
  \caption{\textbf{Comparison on the YCB-Video Dataset.} We report the ADD(-S), and AUC of ADD-S and ADD(-S). Following \cite{xiang2017posecnn}, the symmetric metric is used for all objects in ADD-S while only for symmetric objects in ADD(-S). We highlight the best result in red color, and the second best result in blue color. ``--" denotes unavailable results.}
  \label{tab:result_ycbv}
\end{table}

\subsection{Qualitative Results}

\begin{figure}
  \centering
  \begin{subfigure}[b]{0.48\linewidth}
    \centering
    \includegraphics[width=\linewidth]{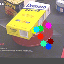}
    \caption{ZebraPose~\cite{su2022zebrapose}}
  \end{subfigure}
  \hfill
  \begin{subfigure}[b]{0.48\linewidth}
    \centering
    \includegraphics[width=\linewidth]{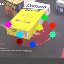}
    \caption{GDR-Net~\cite{wang2021gdr}}
  \end{subfigure}

  \begin{subfigure}[b]{0.48\linewidth}
    \centering
    \includegraphics[width=\linewidth]{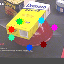}
    \caption{CheckerPose (Ours)}
  \end{subfigure}
  \hfill
  \begin{subfigure}[b]{0.48\linewidth}
    \centering
    \includegraphics[width=\linewidth]{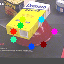}
    \caption{Ground Truth}
  \end{subfigure}
\caption{\textbf{Keypoint localization.} \textbf{(a}) Keypoint locations based on the predicted pose of ZebraPose~\cite{su2022zebrapose}. \textbf{(b)} Keypoint locations based on the pose estimated by GDR-Net~\cite{wang2021gdr}. \textbf{(c)} Keypoint locations output by our network. \textbf{(d)} The ground truth keypoint locations. Considering the symmetry of the bowl, we use the equivalent rotations closest to our prediction to project the keypoints in (a), (b), and (d).}
\label{fig:qual_keypoints}
\end{figure}

In Figure~\ref{fig:qual_keypoints}, we provide localization results of eight keypoints for the occluded and flipped bowl. While our network directly outputs the 2D locations, the results of other dense methods~\cite{su2022zebrapose, wang2021gdr} are computed by projecting the keypoints using the estimated poses.  
Figure~\ref{fig:qual_keypoints} (a) visualizes the reprojections of ZebraPose~\cite{su2022zebrapose}, where the keypoints concentrate on the visible pixels.
Since ZebraPose generates pixel-wise 3D coordinates from the visible regions, it predicts a drastically wrong pose for the severely occluded bowl.
As shown in Figure~\ref{fig:qual_keypoints} (b), the reprojections of GDR-Net~\cite{wang2021gdr} cover the region similar to the ground truth (Figure~\ref{fig:qual_keypoints} (d)). However, the order of the blue keypoint and the red one changes from clockwise to counterclockwise, indicating the bowl is actually faced up. Since GDR-Net is an end-to-end method, it may memorize poses that frequently appear in the training samples. As shown in Figure~\ref{fig:qual_keypoints} (c), our network is capable of localizing the keypoints for the upside-down object with severe occlusion. 
More qualitative results can be found in the Supplementary Material.

\subsection{Runtime Analysis}
We test the running speed on the LM-O dataset. Given a $640 \times 480$ RGB image, we evaluate the speed on a desktop with an Intel 3.30GHz CPU and an NVIDIA GeForce GTX 1080 GPU (8G){, which is reasonable in real-world application}. The FCOS detector~\cite{tian2019fcos} takes 87 ms for each image. The runtime of establishing the dense 3D-2D correspondences by our network is 78 ms. RANSAC/PnP~\cite{lepetit2009epnp} takes only 1 ms to recover pose from the correspondences, while Progressive-X~\cite{barath2019progressive} takes 32 ms. {Under the same testing environment, ZebraPose~\cite{su2022zebrapose} requires 10ms for generating 3D-2D correspondences by CNN and around 350ms to estimate pose using Progressive-X. The overall running time of our method is greatly reduced, because we establish at most 512 candidate 3D-2D correspondences while ZebraPose outputs $128^2$ candidates in the worst case.}

\section{Conclusion}
In this work, we propose a novel way to establish dense correspondences for object pose estimation, by progressively localizing dense 3D keypoints in the input image.
With dense keypoints including occluded and self-occluded ones, we comprehensively explore the available geometry information and enhance the robustness of pose estimation under severe occlusion. We adopt graph neural networks to explicitly model the keypoint interactions, and design a hierarchical binary code representation for the 2D locations. The experiments on LM, LM-O and YCB-V datasets demonstrate that our method achieves state-of-the-art performance of instance-level object pose estimation.  

\vspace{1mm}\noindent{{\bf Acknowledgement.}
We thank all reviewers for valuable comments and suggestions. The work was supported in part by US National Science Foundation Grants 2006665 and 2128350, and by the Defense Advanced Research Projects Agency (DARPA) under Agreement No. HR0011-22-9-0077. 
This work is also supported in part by the SBU/BNL Seed Grant Award. Any opinions, findings, and conclusions, or recommendations expressed in this material are those of the authors and do not necessarily reflect the views of funding agencies.  }

{\small
\bibliographystyle{ieee_fullname}
\bibliography{egbib}
}
\clearpage

\section{Supplemental Material}

\subsection{Hyper-parameters in the Pose Solver}

We use both RANSAC/PnP~\cite{lepetit2009epnp} and Progressive-X~\cite{barath2019progressive} when evaluating the results on the LM dataset~\cite{hinterstoisser2012model}, and we use Progressive-X for LM-O~\cite{brachmann2014learning} and YCB-V~\cite{xiang2017posecnn} datasets. For both pose solvers, we set the threshold of reprojection error as 2 pixels. We run 150 iterations when using RANSAC/PnP and run 400 iterations when using Progressive-X.

\subsection{{Additional Ablation Experiments on LINEMOD Dataset}}

{Theoretically, increasing the number of keypoint $N$ leads to more candidate 3D-2D correspondences and enhances the robustness of pose estimation. In our current implementation, we adopt $k=20$ in EdgeConv following~\cite{wang2019dynamic}, and $N=512$ based on our available computation resources. We also conduct ablation studies of $N$ and $k$ on the LM dataset in Table~\ref{tab:lm_ablation_n_k}, showing that larger $N$ and $k$ help improve the performance. }

\begin{table}[!h]
  \centering
  \small
  \renewcommand{\tabcolsep}{1.5mm}
\vspace{-1mm}
\begin{tabular}{ c | c | c | c | c | c | c }
\toprule
\multirow{2}{*}{$N$} & \multirow{2}{*}{$k$}& \multicolumn{3}{c|}{ADD(-S)} & \multirow{2}{*}{$2^\circ2$cm} & \multirow{2}{*}{$5^\circ5$cm} \\
\cline{3-5}
 &    & 0.02d  &  0.05d  &  0.1d &  &   \\
\midrule
\multirow{3}{*}{128} & 10 & 29.4 & 81.3 & 96.4 & 75.6 & 98.8 \\
                     & 15 & 29.2 & 81.0 & 96.1 & 74.8 & 98.6 \\
                     & 20 & 29.8 & 82.0 & 96.5 & 77.6 & 98.7 \\
\midrule
\multirow{3}{*}{256} & 10 & \textbf{36.0} & 84.2 & 96.8 & 79.1 & 98.8 \\
                     & 15 & 32.0 & 83.4 & 96.8 & 78.1 & 98.8 \\
                     & 20 & 33.6 & 82.9 & 96.4 & 75.8 & 98.8 \\
\midrule
\multirow{3}{*}{512} & 10 & 30.4 & 82.0 & 96.6 & 76.3 & 98.7 \\
                     & 15 & 29.9 & 82.8 & 96.3 & 76.4 & 98.5 \\
                     & 20 & 35.7 & \textbf{84.5} & \textbf{97.1} & \textbf{79.7} & \textbf{98.9} \\
\bottomrule
\end{tabular}
 \caption{\textbf{Ablation Study of $N$ and $k$ on the LM Dataset.}}
 \label{tab:lm_ablation_n_k}
\end{table}

\subsection{Filtering Operation on LM-O and YCB-V}

As discussed in the main paper, we empirically find that for a textureless object $O$ with severe self-occlusions, filtering out the correspondences outside the visible segmentation masks $M_{\mathrm{vis}}$ can improve the pose estimation results. We quantify the self-occlusions of $O$ using $r_{\rm so}(O)$. As a common practice, the visibility of point $P \in O$ from each viewpoint can be determined by checking the intersections between the camera rays and the object mesh. However, this may produce undesired results for our task. For example, the mesh of the bowl in the YCB-V dataset can be treated as a half sphere with very small thickness. When sampling the dense keypoints from the surface, we get keypoints from both outer side and inner side. For the keypoint on the inner side of the bowl, it is considered as easily self-occluded when we use ray intersections to determine the visibility. However, since the bowl is textureless and the thickness of the mesh can be ignored, the keypoint is equivalent to the nearest surface point on the outer side, and should not be considered as easily self-occluded. Considering this issue and the slow computation speed, we instead use Hidden Point Removal (HPR) operator~\cite{katz2007direct} to estimate the proportion $V(P)$ of the viewpoints for which $P$ is visible. For a keypoint with high $V(P)$, it may be consistently misclassified as invisible by the HPR operator, so we ignore the points with $V(P) < 0.2$ estimated by the HPR operator.

We report the value of $r_{\rm so}(O)$ for each object $O$ of the LM-O dataset in Table~\ref{tab:supp_lmo_selfocc}. Since these objects do not have strong textures, we apply the filtering operation during the inference for the objects with $r_{\rm so}(O) \geq 0.5$.

\begin{table}
  \centering
  \renewcommand{\tabcolsep}{6.0mm}
\begin{tabular}{c|c|c}
\toprule
Object & $r_{\rm so}$ & filtering \\
\midrule
ape     & 0.356 & \xmark  \\
can     & 0.650 & \cmark  \\
cat     & 0.584 & \cmark  \\
driller & 0.657 & \cmark  \\
duck    & 0.483 & \xmark  \\
eggbox  & 0.529 & \cmark  \\
glue    & 0.362 & \xmark  \\
holep.  & 0.354 & \xmark  \\
\bottomrule
\end{tabular}
  \caption{\textbf{Quantitative measure $r_{\rm so}$ of the self-occlusions of the objects on LM-O~\cite{brachmann2014learning}.} Since the objects do not have strong textures, for the objects with $r_{\rm so} \geq 0.5$, we apply the filtering operation during inference, \ie, discarding the correspondences outside the visible segmentation masks.}
  \label{tab:supp_lmo_selfocc}
\end{table}

For the objects that requires filtering operation, we report the ADD(-S) metric without filtering in Table~\ref{tab:supp_lmo_wo_filter}. We also report the results of using different segmentation masks to filter the correspondences in Table~\ref{tab:supp_lmo_wo_filter}. Without the filtering operation, the ADD(-S) values decreases for all the objects. Since all the 2D projections should be located within the full segmentation mask $M_{\mathrm{full}}$, using $M_{\mathrm{full}}$ to filter the correspondences aims to discard the wrong predictions outside the object area. However, it does not improve the final estimations consistently, which indicates that we still need to discard more unstable correspondences within the object area.  

\begin{table}
  \centering
  \small
\begin{tabular}{c|c|c|c}
\toprule
Object & w/o Filter & w/ Filter ($M_{\mathrm{full}}$) & w/ Filter ($M_{\mathrm{vis}}$) \\
\midrule
can     & 95.2 & 95.1 & \textbf{95.7}  \\
cat     & 62.0 & 61.3 & \textbf{62.3}  \\
driller & 92.6 & 92.6 & \textbf{93.7}  \\
eggbox  & 68.8 & 69.6 & \textbf{70.0}  \\
\bottomrule
\end{tabular}
  \caption{\textbf{ADD(-S) metrics on LM-O~\cite{brachmann2014learning} w.r.t.~the filtering operation.} ``w/o Filter" denotes using all predicted correspondences to compute the pose. ``w/ Filter ($M_{\mathrm{full}}$)" denotes discarding the correspondences outside the full segmentation mask $M_{\mathrm{full}}$, while ``w/ Filter ($M_{\mathrm{vis}}$)" denotes discarding the correspondences outside the full segmentation mask $M_{\mathrm{vis}}$.}
  \label{tab:supp_lmo_wo_filter}
\end{table}

We report the values of $r_{\rm so}$ for the textureless objects in the YCB-V dataset in Table~\ref{tab:supp_ycbv_selfocc}. According to Table~\ref{tab:supp_ycbv_selfocc}, only one textureless object, \ie, 061\_foam\_brick, requires filtering operation due to severe self-occlusions. 

\begin{table}
  \centering
\begin{tabular}{c|c|c}
\toprule
Object & $r_{\rm so}$ & filtering \\
\midrule
011\_banana              & 0.240 & \xmark  \\
019\_pitcher\_base       & 0.221 & \xmark  \\
024\_bowl                & 0.498 & \xmark  \\
025\_mug                 & 0.108 & \xmark  \\
036\_wood\_block         & 0.438 & \xmark  \\
037\_scissors            & 0.365 & \xmark  \\
051\_large\_clamp        & 0.163 & \xmark  \\
052\_extra\_large\_clamp & 0.138 & \xmark  \\
061\_foam\_brick         & 0.542 & \cmark  \\
\bottomrule
\end{tabular}
  \caption{\textbf{Quantitative measure $r_{\rm so}$ of the self-occlusions of the textureless objects on YCB-V~\cite{xiang2017posecnn}.} For the object with $r_{\rm so} \geq 0.5$, we apply the filtering operation during inference, \ie, discarding the correspondences outside the visible segmentation masks.}
  \label{tab:supp_ycbv_selfocc}
\end{table}

We further report the ADD(-S) metric w.r.t.~the filtering operation for 061\_foam\_brick in Table~\ref{tab:supp_ycbv_wo_filter}. The ADD(-S) of 061\_foam\_brick remains the same without filtering operation or using $M_{\mathrm{full}}$ rather than $M_{\mathrm{vis}}$ in the filtering operation. This observation suggests that the localization of the easily self-occluded regions may become stable after 380,000 training steps. We further investigate the results of 061\_foam\_brick after different training steps in Table~\ref{tab:supp_ycbv_diff_steps}. After 200,000 steps, the ADD(-S) without filtering is inferior to the result of discarding correspondences outside $M_{\mathrm{vis}}$. This observation implies that the localization of the easily self-occluded regions are unstable with fewer training steps. 

\begin{table}
  \centering
  \small
  \renewcommand{\tabcolsep}{1.0mm}
\begin{tabular}{c|c|c|c}
\toprule
Object & w/o Filter & w/ Filter ($M_{\mathrm{full}}$) & w/ Filter ($M_{\mathrm{vis}}$) \\
\midrule
008\_pudding\_box    & 66.4 & 71.0 & \textbf{86.5}  \\
061\_foam\_brick     & \textbf{87.2} & \textbf{87.2} & \textbf{87.2}  \\
\bottomrule
\end{tabular}
  \caption{\textbf{ADD(-S) metrics on YCB-V~\cite{xiang2017posecnn} w.r.t.~the filtering operation.} ``w/o Filter" denotes using all predicted correspondences to compute the pose. ``w/ Filter ($M_{\mathrm{full}}$)" denotes discarding the correspondences outside the full segmentation mask $M_{\mathrm{full}}$, while ``w/ Filter ($M_{\mathrm{vis}}$)" denotes discarding the correspondences outside the full segmentation mask $M_{\mathrm{vis}}$.}
  \label{tab:supp_ycbv_wo_filter}
\end{table}

\begin{table}
  \centering
  \small
\begin{tabular}{c|c|c|c}
\toprule
Steps & w/o Filter & w/ Filter ($M_{\mathrm{full}}$) & w/ Filter ($M_{\mathrm{vis}}$) \\
\midrule
200k     & 86.1 & 85.4 & 86.8 \\
380k     & 87.2 & 87.2 & 87.2 \\
\bottomrule
\end{tabular}
  \caption{\textbf{ADD(-S) metrics of 061\_foam\_brick with different training steps.} ``w/o Filter" denotes using all predicted correspondences to compute the pose. ``w/ Filter ($M_{\mathrm{full}}$)" denotes discarding the correspondences outside the full segmentation mask $M_{\mathrm{full}}$, while ``w/ Filter ($M_{\mathrm{vis}}$)" denotes discarding the correspondences outside the full segmentation mask $M_{\mathrm{vis}}$.}
  \label{tab:supp_ycbv_diff_steps}
\end{table}

\begin{figure}[t]
\begin{center}
   \includegraphics[width=0.6\linewidth]{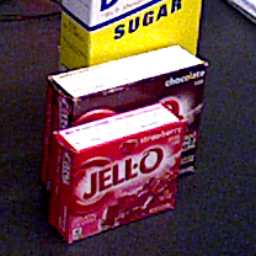}
\end{center}
   \caption{\textbf{Example of test images for 008\_pudding\_box from the YCB-V dataset.} We visualize the zoomed-in RoI based on the detection results. For all test images, 008\_pudding\_box (the brown box) is severely occluded by 009\_gelatin\_box (the red box).}
\label{fig:supp_example_pudding_box}
\end{figure}

Besides textureless objects with severe self-occlusions, we also apply filtering operation on 008\_pudding\_box from the YCB-V dataset. As shown in Figure~\ref{fig:supp_example_pudding_box}, 008\_pudding\_box is severely occluded by 009\_gelatin\_box. We regard 009\_gelatin\_box as a distraction object for the keypoint localization task of 008\_pudding\_box, since these objects share similar appearances, especially the texts (\ie, ``JELL-O"). Such severe occlusions by the same distraction object exist in all the test images of 008\_pudding\_box, and can be automatically detected by checking the object detection results. Thus we discard the correspondences outside $M_{\mathrm{vis}}$ to remove the unstable localization results due to the occlusions by the distraction object. 
We also report the ADD(-S) metric without filtering and using $M_{\mathrm{full}}$ in filtering in Table~\ref{tab:supp_ycbv_wo_filter}. Using either $M_{\mathrm{full}}$ or $M_{\mathrm{vis}}$ to filter the correspondences improve the pose estimation results compared with using all predicted correspondences. This indicates that the filtering operation can remove extreme outliers that are far from 008\_pudding\_box to improve the pose estimation. Using $M_{\mathrm{vis}}$ in the filtering operations obtains better results than $M_{\mathrm{full}}$, which demonstrates that the localization results of the keypoints occluded by the distraction object are not accurate enough for recovering the pose.

\subsection{{Evaluation of  2D-3D Correspondences}}

{The evaluation results in the main paper focus on the final estimated poses. We additionally evaluate the quality of the established dense correspondences before RANSAC. Specifically, for each test sample, we reproject the 3D keypoints by the ground truth pose and compute the mean distance between the reprojection results and predicted 2D locations. For symmetric objects, we use the equivalent rotation closest to our final estimated pose. To obtain the inlier ratio of the estimated correspondences, we regard a keypoint as an inlier if its reprojection error is less than 5 pixels. We compute the average reprojection error and inlier ratio for each object and report the average values over the whole dataset in Table~\ref{tab:supp_eval_corr}.}

\begin{table}
  \centering
  \small
\begin{tabular}{c|c|c|c}
\toprule
Dataset & LM & LM-O & YCB-V \\
\midrule
reprojection error (pixel) & 3.4  & 14.4 & 10.9 \\
inlier ratio (\%)          & 88.4 & 67.8 & 39.6 \\
\bottomrule
\end{tabular}
  \caption{\textbf{Evaluation results of predicted dense correspondences.}}
  \label{tab:supp_eval_corr}
\end{table}

\subsection{BOP Results on LM-O and YCB-V}

We report the performance of our method on LM-O and YCB-Video using the evaluation metrics from BOP challenge~\cite{hodan2020bop} in Table~\ref{tab:supp_result_lmo_bop} and Table~\ref{tab:supp_result_ycbv_bop}, respectively. {We mainly select baselines from officially published work. We also include the results of GDRNPP~\cite{liu2022gdrnpp_bop} for reference, which improves upon GDR-Net~\cite{wang2021gdr} with implementation skills including stronger domain randomization, more powerful detectors, \etc, to compensate for the domain gap between training and test images. Without these implementation skills, our method still achieves comparable performance with the state-of-the-art methods, including the refinement based method~\cite{lipson2022coupled}.}

\begin{table}
  \centering
\begin{tabular}{c|c|c|c|c}
\toprule
Method &   $\mathrm{AR}_{\mathrm{MSPD}}$ & $\mathrm{AR}_{\mathrm{MSSD}}$ & $\mathrm{AR}_{\mathrm{VSD}}$ & $\mathrm{AR}$  \\
\midrule
SurfEmb~\cite{haugaard2022surfemb}  & 85.1 & 64.0 & 49.7 & 66.3 \\
Coupled~\cite{lipson2022coupled}    & 83.1 & 63.3 & 50.1 & 65.5 \\
Zebra~\cite{su2022zebrapose}        & \textcolor{blue}{88.0} & \textcolor{blue}{72.1} & \textcolor{red}{55.2} & \textcolor{red}{71.8} \\
NCF~\cite{huang2022neural}          & --   & --   & --   & 63.2 \\
PFA~\cite{hu2022perspective}        & 83.7 & 66.1 & 52.3 & 67.4 \\
CRT-6D~\cite{castro2023crt}         & 83.7 & 64.0 & 50.4 & 66.0 \\
GDRNPP~\cite{liu2022gdrnpp_bop}     & \textcolor{red}{88.7} & 70.1 & \textcolor{blue}{54.9} & \textcolor{blue}{71.3} \\
\midrule
\textbf{Ours}                       & 87.3 & \textcolor{red}{72.3} & 53.7 & 71.1 \\
\bottomrule
\end{tabular}
  \caption{\textbf{Results on LM-O dataset under BOP setup~\cite{hodan2020bop}.} The results of Coupled~\cite{lipson2022coupled} and NCF~\cite{huang2022neural} are obtained from the original paper, and the results of other methods are obtained from \url{https://bop.felk.cvut.cz/leaderboards/}. We highlight the best result in red color, and the second best result in blue color. ``--" denotes unavailable results.}
  \label{tab:supp_result_lmo_bop}
\end{table}

\begin{table}
  \centering
\begin{tabular}{c|c|c|c|c}
\toprule
Method &   $\mathrm{AR}_{\mathrm{MSPD}}$ & $\mathrm{AR}_{\mathrm{MSSD}}$ & $\mathrm{AR}_{\mathrm{VSD}}$ & $\mathrm{AR}$  \\
\midrule
SurfEmb~\cite{haugaard2022surfemb}  & 77.3 & 62.0 & 54.8 & 64.7 \\
Coupled~\cite{lipson2022coupled}    & 85.2 & 83.5 & \textcolor{red}{78.3} & \textcolor{blue}{82.4} \\
Zebra~\cite{su2022zebrapose}        & \textcolor{blue}{86.4} & 83.0 & 75.1 & 81.5 \\
NCF~\cite{huang2022neural}          & --   & --   & --   & 77.5 \\
PFA~\cite{hu2022perspective}        & 84.9 & 81.4 & 75.8 & 80.7 \\
SC6D~\cite{cai2022sc6d}             & 80.4 & 79.6 & 69.5 & 76.5 \\
CRT-6D~\cite{castro2023crt}         & 77.4 & 77.6 & 70.6 & 75.2 \\
GDRNPP~\cite{liu2022gdrnpp_bop}     & \textcolor{red}{86.9} & \textcolor{red}{84.6} & \textcolor{blue}{76.0} & \textcolor{red}{82.5} \\
\midrule
\textbf{Ours}                       & 85.3 & \textcolor{blue}{84.4}   & 70.7   & 80.1    \\
\bottomrule
\end{tabular}
  \caption{\textbf{Results on YCB-Video dataset under BOP setup~\cite{hodan2020bop}.} The results of Coupled~\cite{lipson2022coupled} and NCF~\cite{huang2022neural} are obtained from the original paper, and the results of other methods are obtained from \url{https://bop.felk.cvut.cz/leaderboards/}. We highlight the best result in red color, and the second best result in blue color. ``--" denotes unavailable results.}
  \label{tab:supp_result_ycbv_bop}
\end{table}

\begin{table*}
    \centering
    \begin{tabular}{c|c|c|c|c|c|c|c}
    \toprule
    Method & SegDriven\cite{hu2019segmentation} & S.\@Stage\cite{hu2020single} & RePose~\cite{iwase2021repose}   & GDR~\cite{wang2021gdr} & Zebra~\cite{su2022zebrapose} & DProST~\cite{park2021dprost} & \textbf{Ours} \\
    \midrule
    002\_master\_chef\_can    & 33.0 & - & - & 41.5 & \textbf{62.6} & - & 45.9 \\
    003\_cracker\_box         & 44.6 & - & - & 83.2 & \textbf{98.5} & - & 94.2 \\
    004\_sugar\_box           & 75.6 & - & - & 91.5 & 96.3 & - & \textbf{98.3} \\
    005\_tomato\_soup\_can    & 40.8 & - & - & 65.9 & 80.5 & - & \textbf{83.2} \\
    006\_mustard\_bottle      & 70.6 & - & - & 90.2 & \textbf{100.0} & - & 99.2 \\
    007\_tuna\_fish\_can      & 18.1 & - & - & 44.2 & 70.5 & - & \textbf{88.9} \\
    008\_pudding\_box         & 12.2 & - & - &  2.8 & \textbf{99.5} & - & 86.5 \\
    009\_gelatin\_box         & 59.4 & - & - & 61.7 & \textbf{97.2} & - & 86.0 \\
    010\_potted\_meat\_can    & 33.3 & - & - & 64.9 & \textbf{76.9} & - & 70.0 \\
    011\_banana               & 16.6 & - & - & 64.1 & 71.2 & - & \textbf{96.0} \\
    019\_pitcher\_base        & 90.0 & - & - & 99.0 & \textbf{100.0} & - & \textbf{100.0} \\
    021\_bleach\_cleanser     & 70.9 & - & - & 73.8 & 75.9 & - & \textbf{89.8} \\
    024\_bowl*                & 30.5 & - & - & 37.7 & 18.5 & - & \textbf{68.0} \\
    025\_mug                  & 40.7 & - & - & 61.5 & 77.5 & - & \textbf{89.0} \\
    035\_power\_drill         & 63.5 & - & - & 78.5 & \textbf{97.4} & - & 95.9 \\
    036\_wood\_block*         & 27.7 & - & - & 59.5 & \textbf{87.6} & - & 58.7 \\
    037\_scissors             & 17.1 & - & - &  3.9 & \textbf{71.8} & - & 62.4 \\
    040\_large\_marker        &  4.8 & - & - &  7.4 & \textbf{23.3} & - & 18.8 \\
    051\_large\_clamp*        & 25.6 & - & - & 69.8 & 87.6 & - & \textbf{95.4} \\
    052\_extra\_large\_clamp* &  8.8 & - & - & 90.0 & \textbf{98.0} & - & 95.6 \\
    061\_foam\_brick*         & 34.7 & - & - & 71.9 & \textbf{99.3} & - & 87.2 \\
    \midrule
    MEAN                      & 39.0 & 53.9 & 62.1 & 60.1 & 80.5 & 65.1 & \textbf{81.4} \\
    \bottomrule
    \end{tabular}
    \caption{\textbf{Detailed results on YCB-V \cite{xiang2017posecnn} w.r.t.~ADD(-S).} (*) denotes symmetric objects and ``-" denotes unavailable results.}
    \label{tab:supp_ycbv_add}
\end{table*}

\begingroup
\setlength{\tabcolsep}{3pt}
\begin{table*}
    \centering
    \begin{tabular}{c|cc|cc|cc|c|cc}
    \toprule
     Method & \multicolumn{2}{c|}{CosyPose~\cite{labbe2020cosypose}} & \multicolumn{2}{c|}{GDR-Net\cite{wang2021gdr}} & \multicolumn{2}{c|}{ZebraPose\cite{su2022zebrapose}} & DProST~\cite{park2021dprost} &
     \multicolumn{2}{c}{\textbf{Ours}} \\
    \midrule
    \multirow{2}{*}{Metric}   & AUC of & AUC of  & AUC of & AUC of  & AUC of & AUC of  & AUC of & AUC of & AUC of \\
                              & ADD-S  & ADD(-S) & ADD-S  & ADD(-S) & ADD-S  & ADD(-S) & ADD(-S) & ADD-S  & ADD(-S) \\
    \hline
    002\_master\_chef\_can    & - & - & 96.3 & 65.2 & 93.7 & 75.4 & - & 87.5 & 67.7 \\
    003\_cracker\_box         & - & - & 97.0 & 88.8 & 93.0 & 87.8 & - & 93.2 & 86.7 \\
    004\_sugar\_box           & - & - & 98.9 & 95.0 & 95.1 & 90.9 & - & 95.9 & 91.7 \\
    005\_tomato\_soup\_can    & - & - & 96.5 & 91.9 & 94.4 & 90.1 & - & 94.0 & 89.9 \\
    006\_mustard\_bottle      & - & - & 100.0 & 92.8 & 96.0 & 92.6 & - & 95.7 & 90.9 \\
    007\_tuna\_fish\_can      & - & - & 99.4 & 94.2 & 96.9 & 92.6 & - & 97.5 & 94.4 \\
    008\_pudding\_box         & - & - & 64.6 & 44.7 & 97.2 & 95.3 & - & 94.9 & 91.5 \\
    009\_gelatin\_box         & - & - & 97.1 & 92.5 & 96.8 & 94.8 & - & 96.1 & 93.4 \\
    010\_potted\_meat\_can    & - & - & 86.0 & 80.2 & 91.7 & 83.6 & - & 86.4 & 80.4 \\
    011\_banana               & - & - & 96.3 & 85.8 & 92.6 & 84.6 & - & 95.7 & 90.1 \\
    019\_pitcher\_base        & - & - & 99.9 & 98.5 & 96.4 & 93.4 & - & 95.8 & 91.9 \\
    021\_bleach\_cleanser     & - & - & 94.2 & 84.3 & 89.5 & 80.0 & - & 90.6 & 83.2 \\
    024\_bowl*                & - & - & 85.7 & 85.7 & 37.1 & 37.1 & - & 82.5 & 82.5 \\
    025\_mug                  & - & - & 99.6 & 94.0 & 96.1 & 90.8 & - & 96.9 & 92.7 \\
    035\_power\_drill         & - & - & 97.5 & 90.1 & 95.0 & 89.7 & - & 94.7 & 88.8 \\
    036\_wood\_block*         & - & - & 82.5 & 82.5 & 84.5 & 84.5 & - & 68.3 & 68.3 \\
    037\_scissors             & - & - & 63.8 & 49.5 & 92.5 & 84.5 & - & 91.7 & 81.6 \\
    040\_large\_marker        & - & - & 88.0 & 76.1 & 80.4 & 69.5 & - & 83.3 & 72.3 \\
    051\_large\_clamp*        & - & - & 89.3 & 89.3 & 85.6 & 85.6 & - & 90.0 & 90.0 \\
    052\_extra\_large\_clamp* & - & - & 93.5 & 93.5 & 92.5 & 92.5 & - & 91.6 & 91.6 \\
    061\_foam\_brick*         & - & - & 96.9 & 96.9 & 95.3 & 95.3 & - & 94.1 & 94.1 \\
    \midrule
    MEAN                      & 89.8 & 84.5 & 91.6 & 84.3 & 90.1 & 85.3 & 77.4 & 91.3 & 86.4 \\
    \bottomrule
    \end{tabular}
    \caption{\textbf{Detailed results on YCB-V \cite{xiang2017posecnn} w.r.t.~AUC of ADD-S and ADD(-S).} As in \cite{xiang2017posecnn}, symmetric metric is used for all objects in ADD-S while only for symmetric objects in ADD(-S). (*) denotes symmetric objects.}
    \label{tab:supp_ycbv_auc}
\end{table*}
\endgroup

\subsection{Detailed Results of YCB-V}

We report the detailed evaluation metrics of each object on YCB-V dataset~\cite{xiang2017posecnn} in Table~\ref{tab:supp_ycbv_add} and Table~\ref{tab:supp_ycbv_auc}. Our method outperforms previous methods w.r.t.~ADD(-S) and AUC of ADD(-S), and achieves comparable performance with state of the art w.r.t. AUC of ADD-S.

\subsection{Qualitative Results}

\begin{figure*}[!t]
    \centering
    \includegraphics[width=0.8\linewidth]{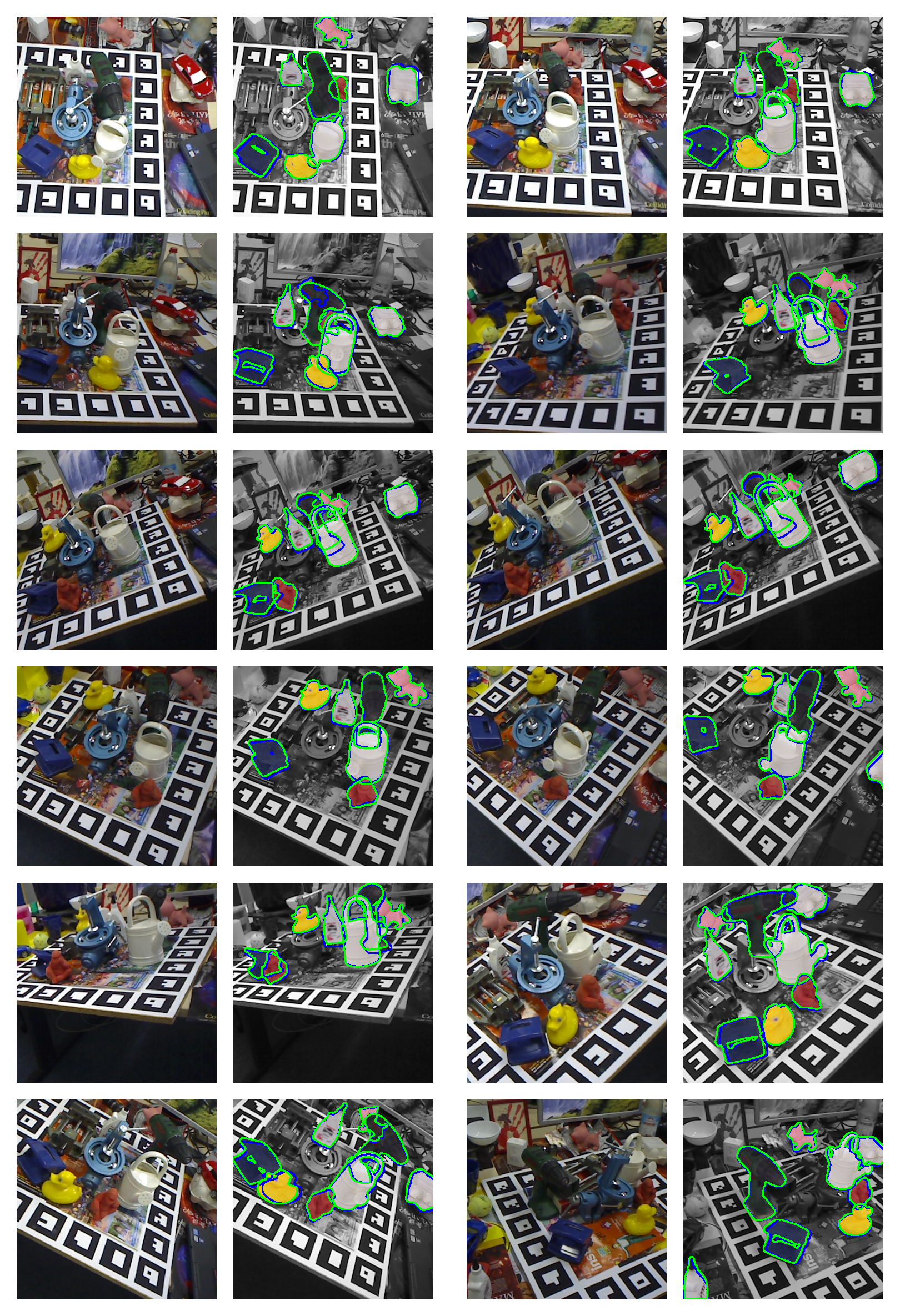}
    \caption{\textbf{Qualitative results on the LM-O dataset.} For each image on the left, we visualize the 6D pose by rendering the 3D CAD models and highlighting the contours on the right. Blue color denotes ground truth and green color denotes the prediction from CheckerPose.}
    \label{fig:supp_qual_lmo}
\end{figure*}

\begin{figure*}[!t]
    \centering
    \includegraphics[width=\linewidth]{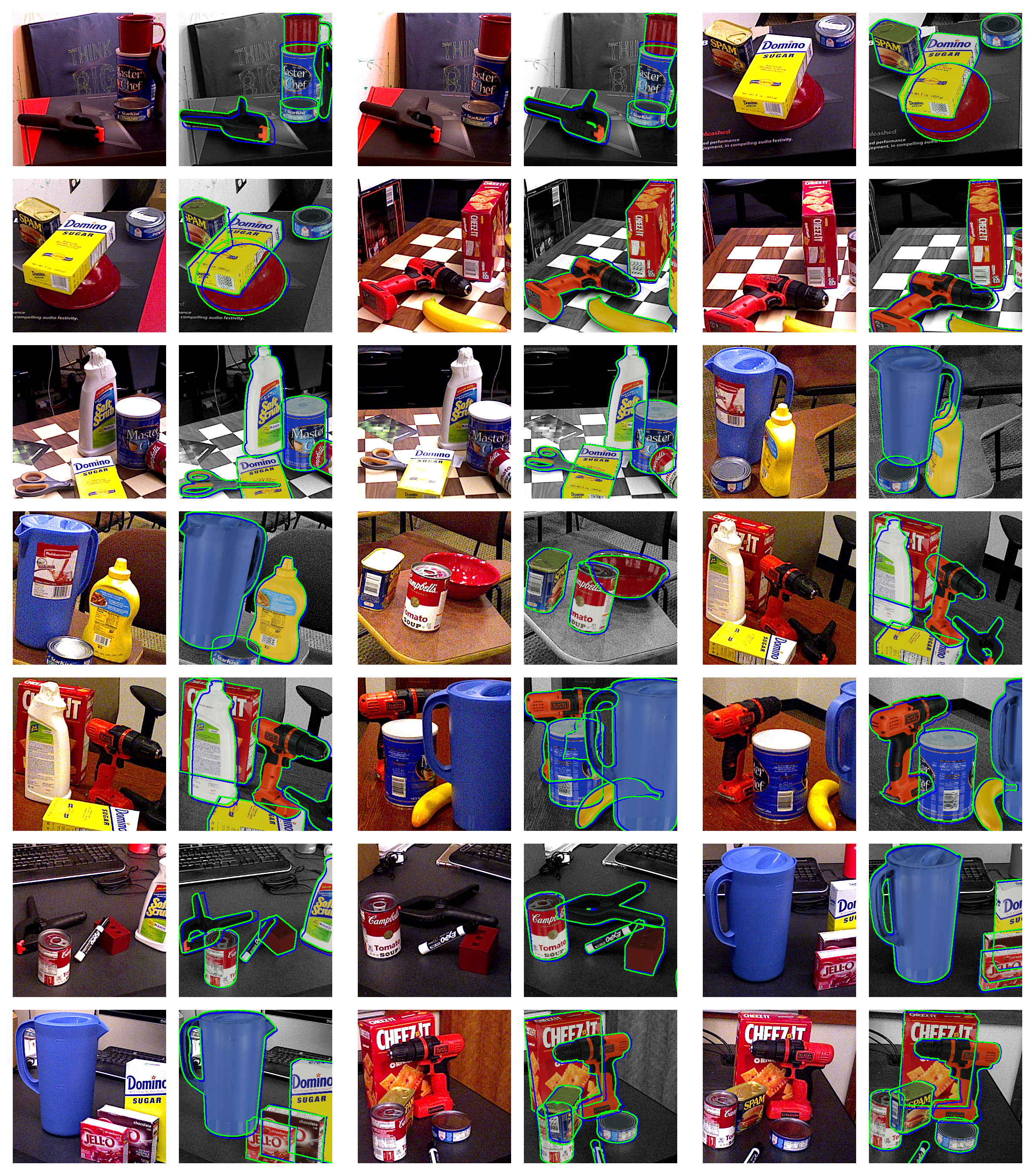}
    \caption{\textbf{Qualitative results on the YCB-V dataset.} For each image on the left, we visualize the 6D pose by rendering the 3D CAD models and highlighting the contours on the right. Blue color denotes ground truth and green color denotes the prediction from CheckerPose.}
    \label{fig:supp_qual_ycbv}
\end{figure*}

We provide additional qualitative results for LM-O \cite{brachmann2014learning} and YCB-V \cite{xiang2017posecnn} in Figure~\ref{fig:supp_qual_lmo} and Figure~\ref{fig:supp_qual_ycbv}, respectively. We render the 3D CAD model based on the predictions of CheckerPose, and highlight the contour in green. We also highlight the ground truth contour in blue. For better visualization, we crop the images and we also show the original input image on the left for LM-O and YCB-V. 

\begin{figure*}[!t]
    \centering
    \includegraphics[width=0.6\linewidth]{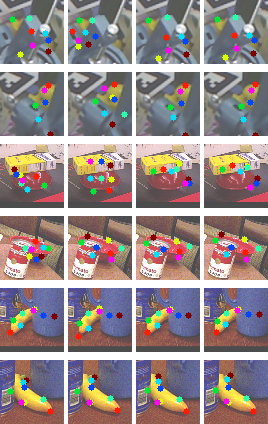}
    \caption{\textbf{Visualization of keypoint localization.} Each column visualizes the keypoint location results of ZebraPose~\cite{su2022zebrapose}, GDR-Net~\cite{wang2021gdr}, our method, and ground truth. While our network directly outputs the 2D locations, the results of other dense methods~\cite{su2022zebrapose, wang2021gdr} are computed by projecting the keypoints using the estimated poses.}
    \label{fig:supp_qual_keypoint}
\end{figure*}

Furthermore, we provide more keypoint localization results of duck, bowl, and banana in Figure~\ref{fig:supp_qual_keypoint}. For better visualization we only plot eight keypoints that are evenly distributed over the object surface. 
While our network directly outputs the 2D locations, the results of other dense methods~\cite{su2022zebrapose, wang2021gdr} are computed by projecting the keypoints using the estimated poses. Considering the symmetry of the bowl, we use the equivalent rotations closest to our prediction to project the keypoints of bowl.

\subsection{{Failure Cases and Future Work}}

\begin{figure*}
  \centering
  \begin{subfigure}[b]{0.15\linewidth}
    \centering
    \includegraphics[width=\linewidth]{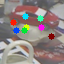}
    \caption{Ground Truth}
  \end{subfigure}
  \hspace{0.02\linewidth}
  \begin{subfigure}[b]{0.15\linewidth}
    \centering
    \includegraphics[width=\linewidth]{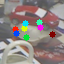}
    \caption{Prediction}
  \end{subfigure}
  \hspace{0.02\linewidth}
  \begin{subfigure}[b]{0.15\linewidth}
    \centering
    \includegraphics[width=\linewidth]{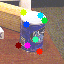}
    \caption{Ground Truth}
  \end{subfigure}
  \hspace{0.02\linewidth}
  \begin{subfigure}[b]{0.15\linewidth}
    \centering
    \includegraphics[width=\linewidth]{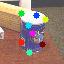}
    \caption{Prediction}
  \end{subfigure}
\caption{\textbf{Failure cases.} We provide the localization results of eight keypoints that are inliers of the estimated poses.}
\label{fig:qual_failure}
\end{figure*}

{We visualize typical failure cases in Figure~\ref{fig:qual_failure}. As shown in Figure~\ref{fig:qual_failure} (a) and (b), the textureless object eggbox from LM-O dataset is severely occluded by a toy car, and a distraction object with similar color also partially appears in the input RoI. As a result, the estimated 2D projections are shifted towards the distraction object. We also present a failure case of objects with textures in Figure~\ref{fig:qual_failure} (c) and (d). The object in interest is 002\_master\_chef\_can from YCB-V dataset, which is geometrically symmetric. Though the texture is almost symmetric as well, the barcode only appears on one side of the object, which causes the asymmetry. For the given input RoI, the keypoints are localized in the opposite directions, w.r.t. the central axis.}

{To improve the localization results, one future direction is the selection of 3D keypoints. Since we adopt farthest point sampling algorithm to obtain evenly distributed keypoints, we ignore other factors to make the keypoints more representative. For example, the issue of 002\_master\_chef\_can may be solved by sampling more keypoints in the barcode area.
Besides, no positional encoding~\cite{mildenhall2020nerf, tancik2020fourfeat} is leveraged in graph feature aggregation and image feature fusion operations. Such encoding can provide additional cues for textureless regions. In future, we will explore the positional encoding to enhance the keypoint localization process.}

\end{document}